\documentclass[journal]{IEEEtran}

\usepackage{cite}

\ifCLASSINFOpdf
  \usepackage[pdftex]{graphicx}
\else
  \usepackage[dvips]{graphicx}
\fi
\usepackage{amsmath}
\usepackage{algorithmic}
\usepackage[linesnumbered,ruled]{algorithm2e}

\usepackage{array}

\ifCLASSOPTIONcompsoc
  \usepackage[caption=false,font=normalsize,labelfont=sf,textfont=sf]{subfig}
\else
  \usepackage[caption=false,font=footnotesize]{subfig}
\fi
\usepackage{fixltx2e}

\usepackage{stfloats}
\newcolumntype{P}[1]{>{\centering\arraybackslash}p{#1}}

\usepackage{booktabs}

\begin{document}
%
\title{Surrogate-assisted Particle Swarm Optimisation for Evolving Variable-length Transferable Blocks for Image Classification}
%
%
%

\author{Bin Wang,~\IEEEmembership{Student Member,~IEEE,}
		Bing Xue,~\IEEEmembership{Member,~IEEE,}
        Mengjie Zhang,~\IEEEmembership{Fellow,~IEEE,}
		\thanks{B. Wang, B. Xue, and M. Zhang are with the School of Engineering and Computer Science, Victoria University of Wellington, PO Box 600, Wellington 6140, New Zealand (e-mails: bin.wang@ecs.vuw.ac.nz; bing.xue@ecs.vuw.ac.nz; and mengjie.zhang@ecs.vuw.ac.nz)}
}

%
%

\markboth{Journal of \LaTeX\ Class Files,~Vol.~XX, No.~XX, July~2020}%
{Shell \MakeLowercase{\textit{et al.}}: Bare Demo of IEEEtran.cls for IEEE Journals}
%



\maketitle

\begin{abstract}
Deep convolutional neural networks have demonstrated promising performance on image classification tasks, but the manual design process becomes more and more complex due to the fast depth growth and the increasingly complex topologies of convolutional neural networks. As a result, neural architecture search has emerged to automatically design convolutional neural networks that outperform handcrafted counterparts. However, the computational cost is immense, e.g. 22,400 GPU-days and 2,000 GPU-days for two outstanding neural architecture search works named NAS and NASNet, respectively, which motivates this work. A new effective and efficient surrogate-assisted particle swarm optimisation algorithm is proposed to automatically evolve convolutional neural networks. This is achieved by proposing a novel surrogate model, a new method of creating a surrogate dataset and a new encoding strategy to encode variable-length blocks of convolutional neural networks, all of which are integrated into a particle swarm optimisation algorithm to form the proposed method. The proposed method shows its effectiveness by achieving competitive error rates of 3.49\% on the CIFAR-10 dataset, 18.49\% on the CIFAR-100 dataset, and 1.82\% on the SVHN dataset. The convolutional neural network blocks are efficiently learned by the proposed method from CIFAR-10 within 3 GPU-days due to the acceleration achieved by the surrogate model and the surrogate dataset to avoid the training of 80.1\% of convolutional neural network blocks represented by the particles. Without any further search, the evolved blocks from CIFAR-10 can be successfully transferred to CIFAR-100 and SVHN, which exhibits the transferability of the block learned by the proposed method. 
\end{abstract}


%
\IEEEpeerreviewmaketitle

\section{Introduction}
%
%
%
%

\IEEEPARstart{C}{onvolutional} Neural Networks (CNNs) have shown promising performance in tackling image classification tasks \cite{zhao2020visual} \cite{zhao2019object}, and the state-of-the-art classification accuracy records have been constantly broken in recent years. One obvious trend has been growing the depth of CNNs to improve the classification accuracy, from only five convolutional layers of AlexNet \cite{krizhevsky2012imagenet} to tens of layers of VGGNet \cite{simonyan2014very} and hundreds of layers of ResNet \cite{he2016deep} and DenseNet \cite{huang2017densely}. Another observation from the recent CNNs is that the shortcut connections \cite{he2016deep} have been introduced to connect the layers that are not next to each other in CNNs, e.g. ResNet \cite{he2016deep}, DenseNet \cite{huang2017densely}, Wide residual networks \cite{zagoruyko2016wide} and PyramidNets \cite{han2017deep}.  The shortcut connections have broken the traditional feed-forward topology of CNNs, but enable more flexible topologies of CNNs. A serious side effect of the depth increase and topology flexibility is that handcrafting CNNs has become much more complex because the optimal depth and optimal topology are extremely difficult to be found due to the indefinite search space comprised of the depth and the topology of CNNs. Apart from the difficulty, the manual design process requires high-level expertise in both CNNs and the datasets, which is also time-consuming because every design trial has to be trained by the stochastic gradient descent (SGD) algorithm and the training process is slow, especially for deep CNNs.

Consequently, a research area of automatically searching for optimal CNNs has been surging in recent years. Two machine learning techniques have been widely used in this area --- reinforcement learning (RL) and evolutionary computation (EC). For example, the researches \cite{zoph2016neural} \cite{brock2017smash} \cite{zoph2018learning} have demonstrated that the CNNs designed automatically by RL methods can outperform the state-of-the-art handcrafted CNNs. The counterpart of using EC methods to automatically evolve CNNs has also achieved promising performance similar to the RL methods, e.g. \cite{xie2017genetic}, \cite{wang2018hybrid} \cite{real2018regularized} and \cite{wang2019particle}. However, most of the methods obtained good CNNs via evaluating a large number of CNNs, which requires very expensive computational cost for training CNNs. For example, \cite{zoph2016neural} and \cite{zoph2018learning} achieved the state-of-the-art classification accuracy with the computational cost of 22,400 GPU-days and 2,000 GPU-days, respectively. They ran the experiments on hundreds of GPUs to acquire good CNNs within a reasonable time-frame, but most of the researchers or practitioners do not have the luxurious computing resource. In this paper, a surrogate-assisted EC method will be proposed to significantly mitigate the expensive computational cost issue. 

The proposed method is mainly inspired by NasNet \cite{zoph2018learning} and DenseNet \cite{huang2017densely}. NasNet established a new approach towards the efficient search of CNNs by seeking optimal blocks instead of whole CNN architectures. Since the block is much smaller than the whole CNN architecture, the computational cost of training the block is much lower, which, therefore, can accelerate the automatic search process. NasNet also found that the optimal blocks learned from one dataset could be transferred to another dataset. This paper will adopt the strategy of evolving single blocks and explore the transferability of the evolved blocks. Another motivation is from DenseNet. DenseNet has demonstrated the performance improvement by building densely-connected CNNs, but it uses a fixed hyper-parameter called growth rate for each layer in a dense block, which might not be an optimal solution. Hence, the paper will propose an EC approach to exploring various growth rates for each layer. Particle Swarm Optimisation (PSO) will be used as the EC algorithm because PSO is a relatively simple EC algorithm, which is computationally inexpensive and effective for optimising a wide range of functions \cite{kennedy1995particle} \cite{eberhart1995new} \cite{shi1998modified}. 

\textbf{Goals:} The overall goal of this paper is to propose a new surrogate-assisted PSO method to efficiently and effectively evolve transferable blocks with the highlights of a novel surrogate method and the transferability of the evolved block. 
The goal will be achieved by accomplishing the following tasks: 
\begin{itemize}
	\item A novel surrogate model will be proposed to predict the result of performance comparison between two CNNs to avoid the expensive computational cost of training CNNs. The proposed method will transform the performance estimation of CNNs to a simple binary classification task. Support vector machine (SVM) \cite{cortes1995support} \cite{chang2011libsvm} is chosen to solve the classification task. 
	\item In-depth analysis and visualisation will be performed to verify the reliability of the proposed surrogate model. Firstly, the pattern of the data to train the surrogate model will be discovered and visualised. Secondly, the surrogate model will be evaluated on different feature combinations of the data to figure out the best feature combinations for the surrogate model. Finally, the performance of the surrogate model during the evolutionary process will be analysed. 
	\item A new method of creating a surrogate dataset will be proposed. The surrogate dataset is sampled from the original dataset by reducing the sample size and the image resolution to reduce the computational cost. 
	\item A new surrogate-assisted PSO method will be proposed by integrating the surrogate model and surrogate dataset into PSO to automatically evolve CNNs. The surrogate model will filter out the unnecessary evaluations of underperformed particles to accelerate the evolution. Moreover, the analysis and visualisation of the evolutionary process will be done to gain deeper insights into the convergence of the proposed method.
	\item An encoding strategy being able to explore various growth rates for each layer in variable-length blocks will be proposed. Further analysis will also be done to figure out the growth rates that are preferred by different layers in the block. 
\end{itemize}

The remainder of this paper is organised as follows. Section \ref{SSS:effpnet_background} introduces the essential background to understand the proposed method, and Section \ref{SSS:effpnet_method} describes the details of the proposed method. The experiments are designed and illustrated in Section \ref{SSS:effpnet_experiment}, and the experimental results are presented and analysed in Section \ref{SSS:effpnet_result}. In the end, the conclusions are made and the future works are envisaged in Section \ref{SSS:effpnet_conclusion}. 


\section{Background and Related Works}\label{SSS:effpnet_background}



\subsection{DenseNet} \label{SSS:effpnet_background_densenet}

\begin{figure}[ht]
	\centering
	\includegraphics[width=\linewidth]{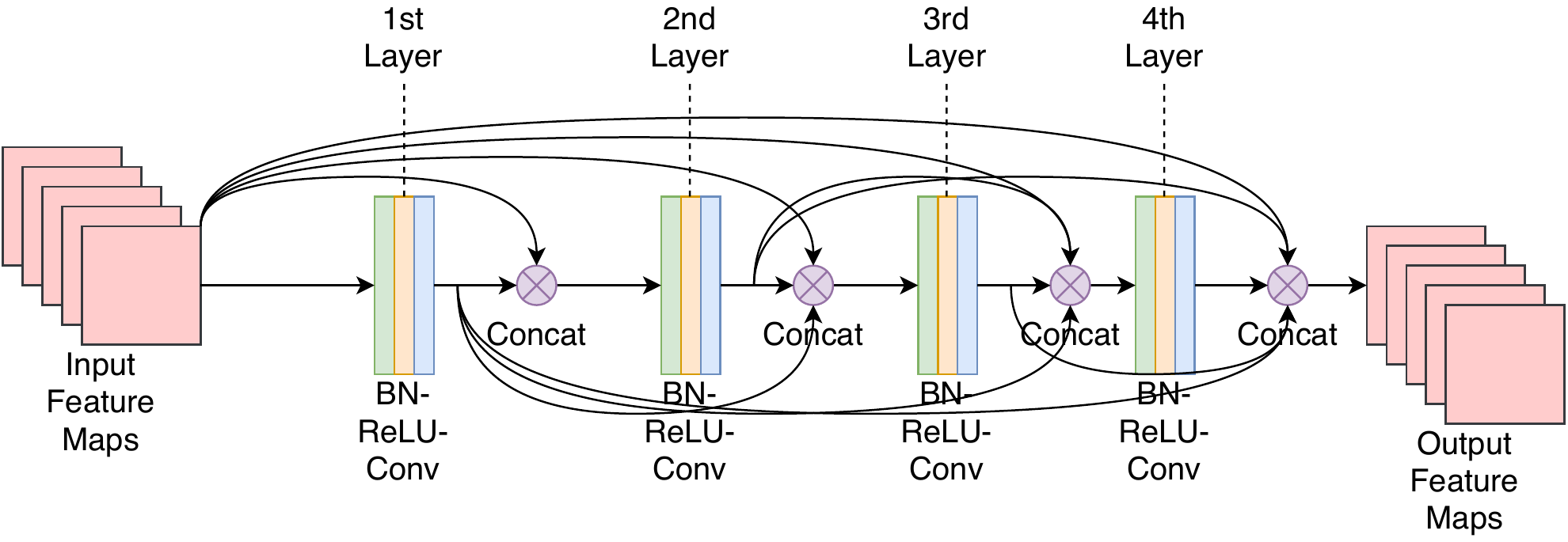}
	\caption{An example of a dense block comprised of four layers.}
	\label{fig:densenet_denseblock}
\end{figure}

Since the dense block is the most fundamental unit in DenseNet \cite{huang2017densely} and will be used in this paper, the details of the dense block are explained first. Fig. \ref{fig:densenet_denseblock} illustrates a dense block of four layers. From the left of the figure, the input is a set of feature maps that are extracted by a convolutional layer \cite{krizhevsky2012imagenet}. The first layer takes the input and produces a set of output feature maps. Instead of passing the output feature maps to the next layer as the input feature maps in other CNNs such as VGGNet \cite{simonyan2014very} and Fitnets \cite{romero2014fitnets}, the output feature maps from the first layer are concatenated with the input feature maps of the first layer to form the input of the second layer. The same strategy applies to the following layers as well. In general, suppose the input feature maps of the dense block are considered as the output feature maps of the layer $0$, the input of the $l_{th}$ layer comes from the concatenation of all the output feature maps from the layers of $0,1...l-1$. In the end, the output feature maps of the dense block are obtained by concatenating the output feature maps of all the layers in the block and the input feature maps of the dense block.

A layer in the dense block is actually a composite layer, which consists of three consecutive layers --- a batch normalisation (BN) layer \cite{ioffe2015batch}, a rectified linear unit (ReLU) layer and a convolutional layer with $3\times3$ filters. The composite layers are written as BN-ReLU-Conv in Fig. \ref{fig:densenet_denseblock}. 


\begin{equation}\label{eq:densenet_growth_rate}
n_{l+1} = \sum_{0}^{l}k_{i}
\end{equation}

There is a key hyper-parameter for the composite layer in the dense block, which is the growth rate \cite{huang2017densely}. The definition of the growth rate is the number of feature maps that the composite layer produces. Assume the number of output feature maps of the $l_{th}$ layer is $k_{l}$, the number of input feature maps of the $(l+1)_{th}$ layer, i.e. $n_{l+1}$, can be derived from Equation (\ref{eq:densenet_growth_rate}), where $k_{0}$ is the number of input feature maps of the dense block. In the experiments of DenseNet paper \cite{huang2017densely}, it takes the approach of having the same growth rate $r$ for all of the layers in the same dense block. Therefore, Equation (\ref{eq:densenet_growth_rate}) can be transformed as $n_{l+1}=k_{0}+r \times l$. 

\begin{figure}[ht]
	\centering
	\includegraphics[width=\linewidth]{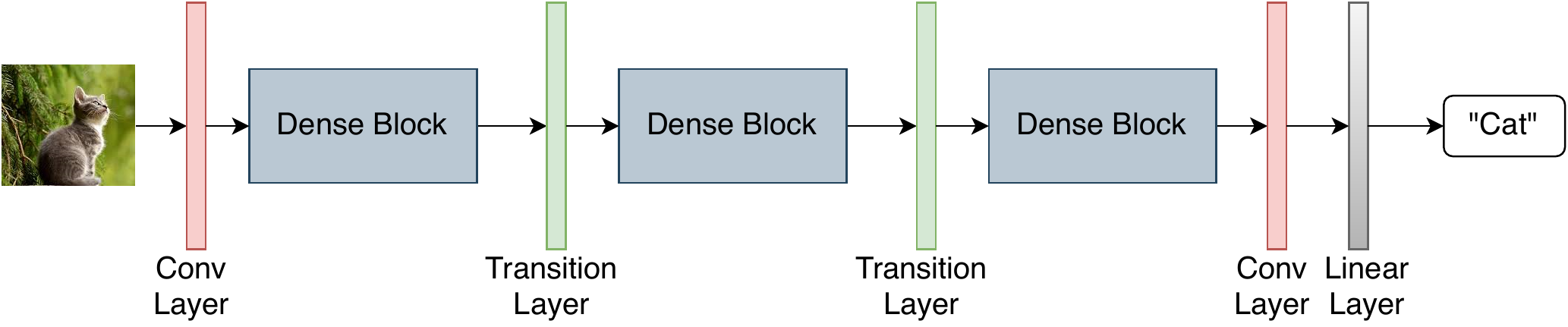}
	\caption{An example of a DenseNet architecture.}
	\label{fig:densenet_densenet}
\end{figure}

DenseNet is primarily composed of dense blocks, and the dense blocks are connected by the transition layers. Fig. \ref{fig:densenet_densenet} exhibits an example of a DenseNet. The input of DenseNet is the image that needs to be classified.  The image is passed to a convolutional layer to produce the feature maps, which are taken by the first dense block as its input feature maps. The dense block produces a set of output features maps, which are processed by the transition layer to generate the input feature maps for the second dense block. By following the same pattern, the third dense block is added. After the last dense block, which is the third dense block in Fig. \ref{fig:densenet_densenet}, a convolutional layer and a linear layer are added consecutively. The final output of the linear layer is used to predict the label of the input image. 

The transition layer is also a composite layer, which is made up of three consecutive layers --- a BN layer, a convolutional layer and a pooling layer. The major functionality of the transition layer in DenseNet is to perform down-sampling to reduce the size of the feature maps, which plays a vital role in CNNs. In the experiments of the DenseNet paper \cite{huang2017densely}, a convolutional filter of $1\times1$ with a stride of 1 is set for the convolutional layer, and a $2\times2$ average pooling layer with a stride of 2 is used as the pooling layer. There is also a hyper-parameter for the transition layer, which is the compression rate $\theta$. Suppose the number of output feature maps of the $l_{th}$ dense block is $n_{l}$, the number of output feature maps produced by the transition layer after the $l_{th}$ dense block is $\theta \times n_{l}$, where $0 < \theta < 1$. In the experiments of DenseNet paper \cite{huang2017densely}, 0.5 is used the $\theta$ value, so after the transition layer, both the number of feature maps and the feature map size are halved. 

\subsection{Particle Swarm Optimisation} \label{SSS:effpnet_background_pso}

\begin{equation}\label{eq:pso_update_v}
v_{id}(t+1) = w v_{id}(t) + c_{1} \epsilon_{1} (p_{id} - x_{id}(t)) + c_{2} \epsilon_{2} (p_{gd} - x_{id}(t))
\end{equation}

\begin{equation}\label{eq:pso_update_x}
x_{id}(t+1) = x_{id}(t) + v_{id}(t+1)
\end{equation}

Particle Swarm Optimisation (PSO) \cite{kennedy1995particle} \cite{bratton2007defining} is essentially inspired by behaviours like bird flocking, fish schooling and swarm theory. Similar to genetic algorithm (GA), PSO initialises the population with random solutions, which are called particles. However, additional to GA, PSO also assigns a random velocity to every particle. The particles fly through the hyperspace only based on primitive mathematical operators shown in Equations (\ref{eq:pso_update_v}) and (\ref{eq:pso_update_x}), called PSO update equations. Both the particle's position and its velocity are represented by vectors. In the update equations, $x_{id}$ means the $d_{th}$ dimension of the $i_{th}$ particle's position, $v_{id}$ represents the $d_{th}$ dimension of the $i_{th}$ particle's velocity, $p_{id}$ is the particle's best-so-far solution at the $d_{th}$ dimension, and $p_{gd}$ keeps the best-so-far solution at the $d_{th}$ dimension across all neighbours of the particle. $(t)$ indicates the current state of the position or velocity, while $(t+1)$ denotes the next state. Apart from the particle's values, there are also several parameters in the update equations. Both $\epsilon_{1}$ and $\epsilon_{2}$ are random values between 0 and 1, $c_{1}$ and $c_{2}$ are two \textit{acceleration coefficients}, and $w$ is called \textit{inertia weight}. 
$w$ is a value between 0 to 1, which is used to avoid the explosion of the velocity.

\subsection{Related Work}

In recent years, RL methods have been widely investigated to automatically design CNNs following the successful research \cite{zoph2016neural} named Neural Architecture Search (NAS). NAS uses a recurrent network as a controller to generate variable-length strings, which represent CNN architectures, and RL optimises the controller to find the optimal CNNs. NAS achieved state-of-the-art CNNs by consuming the computing resource of 22,400 GPU-days. As the computational cost is too expensive, several excellent RL methods have been proposed to reduce the computational cost. For example, NASNet \cite{zoph2018learning} adopts a similar search strategy as NAS, but only searching for a single CNN block, which successfully reduced the computational cost to 2,000 GPU-days. PNASNet \cite{liu2018progressive} and BockQNN \cite{zhong2018practical} have further reduced the computational cost, but their classification accuracy was sacrificed comparing to NASNet.  

Another emerging approach to automatically searching for optimal CNNs is EC-based methods. Early research is LS-Evolution \cite{real2017large}, which employs evolutionary algorithms to evolve CNNs. It achieved promising performance by evolving CNNs for more than 2,730 GPU-days. After that, there are a few successful EC-based methods proposed, such as GeNet \cite{xie2017genetic}, CGP-CNN \cite{suganuma2017genetic} and EIGEN \cite{ren2019eigen}, which reduces the computational cost by compromising the classification accuracy comparing to LS-Evolution. Recently, AmoebaNet \cite{real2018regularized} and AECNN \cite{sun2019completely} were proposed. AmoebaNet proposed a regularized evolutionary algorithm to evolve CNNs, which achieved the state-of-the-art CNNs by using EC methods for the first time, but it took 3,150 GPU-days. Conversely, AECNN accelerated the search process by evaluating fewer CNNs, which only consumed 27 GPU-days, but the classification accuracy was compromised comparing to AmoebaNet. 

From the existing work, it can be observed that the trade-off between the classification accuracy and the computational cost is a challenging problem in automatically designing CNNs both for RL methods and EC methods. In this paper, a novel reliable surrogate model will be proposed to assist PSO to reduce the computational cost by maintaining or even increasing the classification accuracy. The main reason of adopting the EC method is that EC methods tend to converge faster in searching for CNN architectures \cite{real2018regularized}. 

\section{The Proposed Method --- EffPNet} \label{SSS:effpnet_method}

In this section, the details of the proposed method, which is referred as \textit{EffPNet} (Efficient PSO Network), will be illustrated and discussed. Since the evolved blocks learned from one dataset are expected to be transferable to other datasets, the evolved blocks are called \textit{transferable blocks}. The overall framework will be outlined in Section \ref{SSS:effpnet_method_framework}, and the encoding strategy will be described in Section \ref{SSS:effpnet_encoding}. The fitness evaluation, the surrogate model, the surrogate dataset, and the evolutionary process with the surrogate-assistance will be detailed in Section \ref{SSS:effpnet_fitness}, \ref{SSS:effpnet_surrogate_model}, \ref{SSS:effpnet_surrogate_dataset} and \ref{SSS:effpnet_evolve}, respectively. Last but not least, the method of stacking the evolved block will be explained in Section \ref{SSS:effpnet_stack}. 

\subsection{The Overall Framework} \label{SSS:effpnet_method_framework}

\begin{figure*}[ht]
	\centering
	\includegraphics[width=\linewidth]{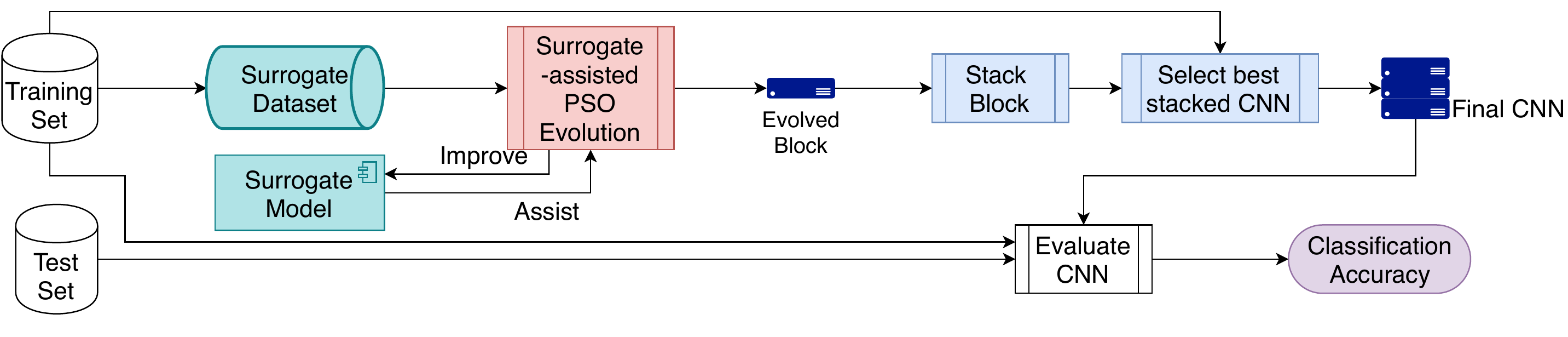}
	\caption{Framework of the proposed surrogate-assisted PSO method.}
	\label{fig:effpnet_framework}
\end{figure*}

The overall framework of the proposed method is outlined in Fig. \ref{fig:effpnet_framework}. First of all, a surrogate dataset is sampled from the training set of the given dataset according to the method detailed in Section \ref{SSS:effpnet_surrogate_dataset}. The surrogate dataset is a small subset of the training set, which will be passed to the evolutionary process. Secondly, the surrogate-assisted PSO method searches for the optimal dense block by only using the surrogate dataset. A surrogate model is trained on the data extracted during the evolutionary process (to be described in Section \ref{SSS:effpnet_surrogate_model}), comprised of the encoded vectors and the corresponding fitness values, which is used to predict the CNN performance instead of training the CNNs by SGD. The details of the surrogate model will be described in Section \ref{SSS:effpnet_surrogate_model}. By combining the surrogate dataset and the surrogate model, the computational cost can be significantly reduced. Thirdly, the evolved dense block is stacked by various numbers of times, e.g. once, twice and three times in Fig. \ref{fig:effpnet_framework}, to produce various CNN architectures as the final candidates. Only the training set is taken to evaluate the stacked CNNs, and the best candidate is selected as the final CNN. In the end, the final CNN is re-trained on the training set and evaluated on the test set. The classification accuracy on the test set will be reported.

\subsection{Encoding Strategy}\label{SSS:effpnet_encoding}

The encoding strategy is designed to encode the hyper-parameters of dense blocks \cite{huang2017densely} with variable lengths. In DenseNet, there are two hyper-parameters --- the number of layers and the growth rate. It uses the same growth rate for all of the layers in a dense block. However, the fixed growth rate of each layer in a dense block is not necessarily optimal, so the proposed method explores dense blocks with various growth rates for each layer besides the number of layers in a dense block in DenseNet \cite{huang2017densely}. 

\begin{figure}[ht]
	\centering
	\includegraphics[width=0.8\linewidth]{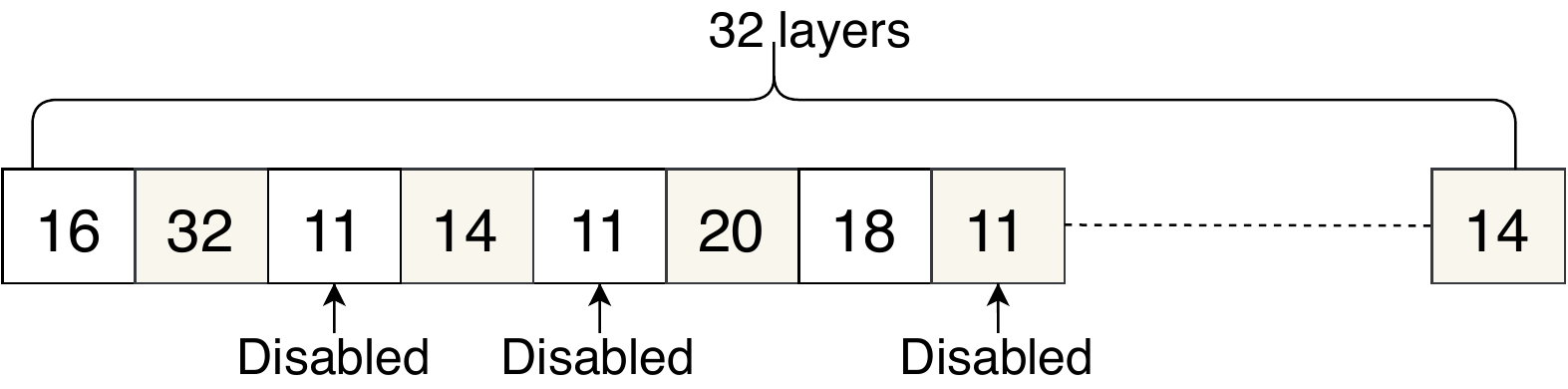}
	\caption{An example of an encoded vector with the maximum number of layers set to 32.}
	\label{fig:effpnet_encoding}
\end{figure}

In order to encode the above dense blocks in each particle, a fixed-length vector shown in Fig. \ref{fig:effpnet_encoding} with the maximum number of layers in a dense block as the dimensionality is proposed to accommodate the hyper-parameters. Each dimension in the vector represents the growth rate of the corresponding layer. A special value is introduced to indicate that the corresponding layer is disabled to achieve variable-length dense blocks. There are three hyper-parameters that need to be defined to accomplish the encoding vector. The first one is the \textit{maximum number of layers} $\mathbf{ml}$ in a dense block. The second one is the range of the growth rates --- the \textit{lower bound} $\mathbf{g_{l}}$ and the \textit{upper bound} $\mathbf{g_u}$. The last is the \textit{special value} $\mathbf{sv}$ to disable a layer in the encoded vector. In the example of Fig. \ref{fig:effpnet_encoding}, $ml$ is set to 32, which is based on the capacity of hardware resource and the complexity of the image dataset. 12 is used for $g_{l}$ based on the experimental experience in DenseNet \cite{huang2017densely} because if the growth rate is too small, it will not be able to capture the features in the output feature maps. 32 is used for $g_{u}$ due to the hardware capacity. After defining the range of the growth rates, $sv$ is defined as the value of $g_{l} - 1$, so the value of each dimension in the vector is a continuous value between 12 and 32 inclusive. 11 is used to represent the disabled layer. 


\subsection{Fitness Evaluation}\label{SSS:effpnet_fitness}

\begin{algorithm}[h]
	\caption{Fitness Evaluation}
	\label{alg:effpnet_fitness}
	\begin{algorithmic}[1]
		\renewcommand{\algorithmicrequire}{\textbf{Input:}}
		\renewcommand{\algorithmicensure}{\textbf{Output:}}
		\newcommand{\algorithmicbreak}{\textbf{break}}
		\newcommand{\BREAK}{\STATE \algorithmicbreak}
		\REQUIRE block $b$, \textit{fitness evaluation dataset} $d$;
		\STATE $acc_{best}, epoch_{best}, epoch, acc \leftarrow$ 0, 0, 0, 0;
		\STATE $losses, acc\_{history} \leftarrow$ empty, empty;
		\STATE $d_{train}, d_{test}\leftarrow$ Randomly split $d$ into 80\% as the training part $d_{train}$ and 20\% as the test part $d_{test}$;
		\WHILE{$acc>=acc_{best}$ \textbf{or} $epoch-epoch_{best}<5$}
		\STATE Apply Adam optimisation \cite{kingma2014adam} to train $b$ on $d_{train}$ \linebreak for one epoch;
		\STATE $losses \leftarrow$ Append training loss to $losses$;
		\STATE $acc\leftarrow$ Evaluate $b$ on $d_{test}$;
		\STATE $acc\_{history} \leftarrow$ Append $acc\leftarrow$ to $acc\_{history}$;
		\IF{$acc > acc_{best}$}
		\STATE $acc_{best}, epoch_{best}\leftarrow$ $acc, epoch$;
		\ENDIF
		\STATE $epoch\leftarrow$ $epoch + 1$;
		\ENDWHILE
		\STATE Save $block, losses, acc\_{history}, acc\_{best}$ as one row \linebreak to a file or database as \textit{block training history};
		\RETURN $acc_{best}$;
	\end{algorithmic}
\end{algorithm}

The details of the fitness evaluation method are illustrated in Algorithm \ref{alg:effpnet_fitness}. The dense block represented by a particle in the PSO swarm is passed to the fitness evaluation function along with the \textit{fitness evaluation dataset}. The \textit{fitness evaluation dataset} can be the whole training dataset or a subset of the training dataset, which is split into the \textit{training part} and the \textit{test part}. The training part is used to train the dense block by SGD. The test part is used to evaluate the trained dense block to produce the classification accuracy, which is used as the fitness value. 

During the above fitness evaluation process, a set of data are collected from the process of training the dense block, which will be used by the surrogate model later in Sections \ref{SSS:effpnet_surrogate_model} and \ref{SSS:effpnet_surrogate_dataset}. During the training
process, the training loss and the classification accuracy on the test part of every epoch are recorded along with the best classification accuracy. These recorded data during the training and the parameters of the dense block are combined and saved as one row of records in a file or a database called \textit{block training history}, which will later be used to construct the data to train the surrogate model in Section \ref{SSS:effpnet_surrogate_model}. 

One important decision made for the fitness evaluation is that Adam optimisation \cite{kingma2014adam} is chosen as the SGD algorithm to train the dense blocks. The main advantage of Adam optimisation is that the learning rate is adjusted based on the training status to optimise the learning rate at the specific epoch for the specific CNN. Due to the adjustment of the learning rate, Adam optimisation demonstrates faster convergence compared with the SGD algorithm \cite{sutskever2013importance} with a fixed learning rate, which can accelerate the fitness evaluation. Besides, the adjustable learning rate can automatically optimise the training process for a CNN, so the classification accuracies obtained for different CNNs will rely less on the pre-specified settings of the SGD method, which provides a fairer comparison between CNNs \cite{wang2020particle}. 



\subsection{Surrogate Model}\label{SSS:effpnet_surrogate_model}

\begin{figure*}[ht]
	\centering
	\includegraphics[width=\linewidth]{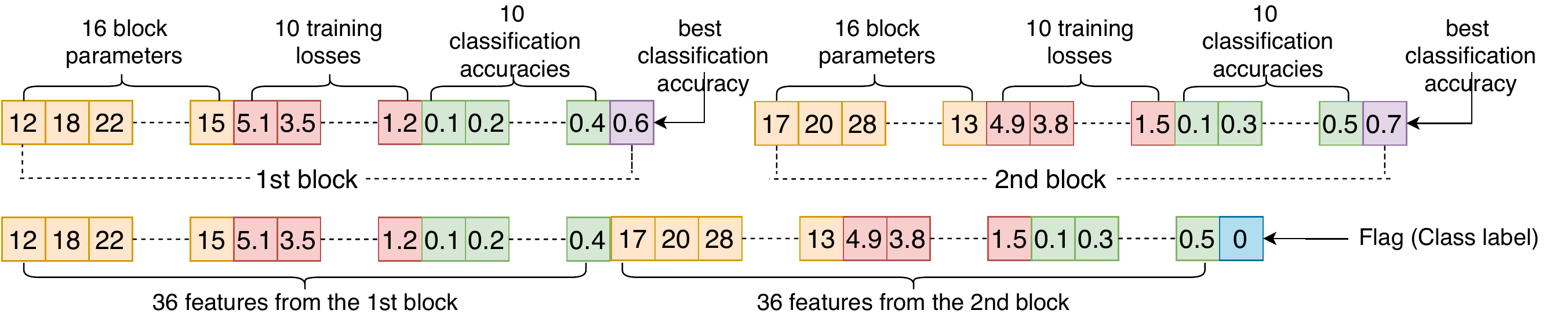}
	\caption{An example of constructing the dataset from the \textit{block training history} to fit the SVM classifier.}
	\label{fig:effpnet-surrogate-dataset-construction}
\end{figure*}

Since PSO only updates the particle's best-so-far solution when the particle's new solution outperforms the best-so-far solution, it is not necessary to acquire the fitness value when the particle's new solution does not perform better. The objective of the proposed surrogate model is to predict whether the CNN represented by the particle's new solution would outperform that represented by the particle's best-so-far solution. First of all, the data used to train the surrogate model need to be constructed from the \textit{block training history} obtained during the fitness evaluation, which transforms the direct prediction of a CNN's performance to a binary classification task of comparing the performance of a pair of CNNs. The class label will be 1 if the first CNN in the pair outperforms the other. Otherwise, the class label will be 0. Firstly, the features in \textit{block training history} need to be selected to represent the evolved block. Since the parameters of the transferable block are decisive to its performance, the parameters are selected. Both the training losses and the classification accuracies are chosen because the training losses on the training part of the fitness evaluation dataset reflect how well the transferable block is trained and the classification accuracies on the test part of the fitness evaluation dataset represent a kind of generalisation quality. Furthermore, a hyper-parameter named the \textit{feature-cutting epoch} is defined, which is the number of epochs used to extract the training losses and classification accuracies from the \textit{block training history}. The \textit{feature-cutting epoch} should be set based on the training process of the dataset by finding the smallest cutting point of the training process where it would be sufficient to learn a trend of the whole training process. Lastly, the data for training the surrogate model are formed by making pairs of the transferable blocks in the \textit{block training history}. Fig. \ref{fig:effpnet-surrogate-dataset-construction} illustrates an example of how an instance of the surrogate model's training data are constructed from a pair of records to form the binary classification task, each of which represents a CNN and its corresponding training process, in the \textit{block training history}. The \textit{feature-cutting epoch} is set to 10. A pair of transferable blocks are drawn in the first row, each of which has 16 growth rates in the block parameters, the first 10 training losses, the first 10 classification accuracies, and the best classification accuracy extracted from the \textit{block training history}. The second row shows the constructed instance form the above pair. The total number of features of the constructed instance is 72, which includes the 16 growth rates, the training losses of the first 10 epochs and the classification accuracy of the 10 epochs from each of the 2 blocks in the pair, i.e. the formula to calculate the total number of features is $(16+10+10)\times2=72$. In this specific example, the class label of the binary classification task is set to 0 because in the example, the best classification accuracy of the first block is less than that of the second block. Otherwise, the class label would be set to 1. A couple of benefits of doing so are that a much larger number of data instances can be obtained by making pairs of a small number of data to achieve an effect of the data augmentation, and it transforms the performance prediction to a binary classification task to simplify the performance prediction task.


Support Vector Machine (SVM) is chosen as the surrogate model to predict whether one CNN could achieve better performance than the other mainly because of a couple of reasons. Firstly, the data in the \textit{block training history} collected from the fitness evaluation process are limited. SVM do not require a large number of examples as deep neural networks to achieve good performance. Secondly, the computational cost of fitting the SVM model is much less than training a deep learning model, so it can efficiently provide good predictions. The surrogate model is iteratively trained after each generation. 
The scores of 10-fold cross-validation of SVM are obtained to assess the quality of how SVM performs on the transformed binary classification task, which are used as the \textit{scores of the surrogate model}. After that, the SVM model is fit by the training data, which will be used as the surrogate model to perform predictions.

\begin{algorithm}[h]
	\caption{Using the surrogate model to perform predictions}
	\label{alg:effpnet_surrogate_model_prediction}
	\begin{algorithmic}[1]
		\renewcommand{\algorithmicrequire}{\textbf{Input:}}
		\renewcommand{\algorithmicensure}{\textbf{Output:}}
		\newcommand{\algorithmicbreak}{\textbf{break}}
		\newcommand{\BREAK}{\STATE \algorithmicbreak}
		\REQUIRE the first transferable block $b_{1}$, the second transferable block $b_{2}$, \textit{feature-cutting epoch} $c$;
		\IF{$b_{1}$ \textbf{ in } \textit{block training history}}
		\STATE $f_{1} \leftarrow$ Extract features for $b_{1}$;
		\ELSE
		\STATE $f_{1} \leftarrow$ Extract the parameters of $b_{1}$, train $b_{1}$ for $c$ epochs, and then extract features for $b_{1}$;
		\ENDIF
		\STATE $f_{2} \leftarrow$ Follow the above process (lines 1 to 5) to extract the features of $b_{2}$;
		\STATE $f \leftarrow$ Concatenate the features $f_{1}$ and $f_{2}$;
		\STATE $t \leftarrow$  Use surrogate model $m$ to predict the target class based on the combined features $f$;
		\RETURN $t$;
	\end{algorithmic}
\end{algorithm}

The trained surrogate model will be used to predict the comparison result between two transferable blocks. The prediction process is described in Algorithm \ref{alg:effpnet_surrogate_model_prediction}. Three inputs are required in the prediction function --- the two transferable blocks and the \textit{feature-cutting epoch}. The \textit{feature-cutting epoch} is the same as that of the training process. When building the features of an example for the prediction, if the transferable block exists in the \textit{block training history}, the training losses and the classification accuracies will be extracted from the existing data; otherwise, the training losses and the classification accuracies are obtained from training the transferable block for the \textit{feature-cutting epochs}. Following the dataset construction process shown in Fig. \ref{fig:effpnet-surrogate-dataset-construction}, the features are built from the two transferable blocks. The surrogate model takes the built features and predicts the comparison result. 

\subsection{Surrogate Dataset}\label{SSS:effpnet_surrogate_dataset}

To reduce the computational cost of the fitness evaluation, the surrogate dataset, which is a reduced training dataset, is used to evaluate the transferable block instead of the whole training dataset. Two techniques are utilised to form the surrogate dataset. The first one is to downsize the \textit{fitness evaluation dataset} by sampling a small subset from the whole training dataset according to a uniform distribution. By following the uniform distribution, the sample subset has a good chance to keep the characteristics of the whole training dataset, so this could mitigate the sacrifice of the model performance. Another technique is to downsample the original images to smaller images. As the image resolution of the state-of-the-art benchmark dataset could be very high, such as the ImageNet dataset, the downsampled images are recognisable by humans, which means the downsampled images can represent the original images very well. Therefore, the downsampled dataset could be used to evaluate the transferable blocks.  

Two hyper-parameters are needed for the two techniques --- the \textit{data-reduction ratio} $p\%$ and the \textit{downsampling factor} $n$. The \textit{data-reduction ratio} is the percentage of the training dataset that is used as the sample size of the surrogate dataset. The \textit{downsampling factor} is the down-scaling ratio of the image size. For example, the down-scaling scales down the image from the size of $w \times h$ to $\dfrac{w}{n} \times \dfrac{h}{n}$. Through downsizing the \textit{fitness evaluation dataset}, the computational cost of training the transferable block for one epoch is $p\%$ of the original cost. By downsampling the images, the feature map size is reduced to $\dfrac{1}{n \times n}$ of the original size, so both the memory and computational cost required to train the transferable block can be reduced to $\dfrac{1}{n \times n}$ of the original ones. In the combination of these two, the total computational cost of using the surrogate dataset can be reduced to $\dfrac{1}{n \times n} \times p\%$ of the original cost. 


\subsection{Evolving Transferable Blocks by Surrogate-assisted PSO}\label{SSS:effpnet_evolve}

\begin{algorithm}[h]
	\caption{Evolving dense block by surrogate-assisted PSO}
	\label{alg:effpnet_pso}
	\begin{algorithmic}[1]
		\renewcommand{\algorithmicrequire}{\textbf{Input:}}
		\renewcommand{\algorithmicensure}{\textbf{Output:}}
		\newcommand{\algorithmicbreak}{\textbf{break}}
		\newcommand{\BREAK}{\STATE \algorithmicbreak}
		\REQUIRE generations $g$, surrogate model threshold $t$, \textit{feature-cutting epoch} $c$;
		\STATE $p\leftarrow$ Random initialise the particles until the population is filled up;
		\STATE $g_{best}, j\leftarrow$ Empty, 0; 
		\STATE $d_{surrogate} \leftarrow$ Generate the surrogate dataset from the training part of the fitness evaluation dataset according to Section \ref{SSS:effpnet_surrogate_dataset}; 
		\WHILE{$j<g$}
			\STATE Train the surrogate model according to Section \ref{SSS:effpnet_surrogate_model};
		\FOR{particle $i$ in $pop$}
		\STATE $i\leftarrow$ Apply standard PSO operations to update the position of $i$;
		\IF{accuracy of the surrogate model $ \geq t$}
			\STATE $flag \leftarrow$ Use surrogate model $m$ to predict the performance comparison  between the transferable blocks represented by the particle's best-so-far solution and the current solution;
				\IF{$flag = 1$}
				\STATE $fitness\leftarrow$ Use Algorithm \ref{alg:effpnet_fitness} to calculate the fitness for $i$;
				\ELSE
				\STATE $fitness\leftarrow$ 0
				\ENDIF
		\ELSE
			\STATE $fitness\leftarrow$ Use Algorithm \ref{alg:effpnet_fitness} to calculate the fitness for $i$;
		\ENDIF
		\STATE Update the fitness of $i$ by $fitness$;
		\STATE Update the particle's best with the current solution when $fitness > $ \textit{the fitness of the particle's best};
		\ENDFOR
		\STATE $g_{best}\leftarrow$ Update with the best particle among the current $g_{best}$ and $pop$;
		\STATE $j\leftarrow$ $j+1$;
		\ENDWHILE
		\RETURN $g_{best}$;
	\end{algorithmic}
\end{algorithm}

The surrogate model and the surrogate dataset are used to assist the PSO method in the evolutionary process to accelerate the proposed method. The details of the evolutionary process are illustrated in Algorithm \ref{alg:effpnet_pso}. Since the proposed encoding strategy has transformed the variable-length parameters of transferable blocks into a fixed-length vector, the standard PSO operations can be applied. However, there are a couple of points specifically related to the surrogate model and the surrogate dataset that need to be explained. Firstly, two hyper-parameters are defined. The first is the threshold to control the activation of the surrogate model, which is activated only when the mean value of the \textit{scores of the surrogate model} obtained in Section \ref{SSS:effpnet_surrogate_model} is larger than the threshold. The second is the \textit{feature-cutting epoch}, which is the same as the \textit{feature-cutting epoch} in Section \ref{SSS:effpnet_surrogate_model} and Algorithm \ref{alg:effpnet_surrogate_model_prediction}. Furthermore, if the surrogate model is activated, before evaluating the particle's new position, the surrogate model will be used to predict whether the new position of the particle will outperform its best position. The new position is evaluated by the fitness evaluation function only when it is predicted to surpass its best position; otherwise, the fitness value of the new position is set to 0. Therefore, the proposed method can avoid unnecessary fitness evaluation for transferable blocks with poor performance to reduce the computational cost.

\subsection{Stacking and Selecting the Best CNN}\label{SSS:effpnet_stack}

Since the block obtained from Algorithm \ref{alg:effpnet_pso} is learned from the surrogate dataset, it might not be trained sufficiently to capture the complexity of the whole dataset. Therefore, a stacking approach depicted in Algorithm \ref{alg:effpnet_stack} is introduced to enhance the capacity of the final network. The stacking approach is also required to transfer the evolved block to other domains because the capacity of the transferable block might not fit other datasets either. There is a hyper-parameter --- the \textit{maximum number of times} $s_{max}$ to stack the transferable block, which is defined to restrict the maximum capacity of the final network. $s_{max}$ is dependent on the complexity of the whole dataset and the hardware resource. During the stacking process, a set of candidates are generated by stacking the learned block from once to $s_{max}$ times, which are then sent to multiple GPU cards to be evaluated in parallel to speed up the stacking process. After receiving the classification accuracies of all candidates, the best candidate is selected as the final solution. 

\begin{algorithm}[h]
	\caption{Stack and select the best candidate}
	\label{alg:effpnet_stack}
	\begin{algorithmic}[1]
		\renewcommand{\algorithmicrequire}{\textbf{Input:}}
		\renewcommand{\algorithmicensure}{\textbf{Output:}}
		\newcommand{\algorithmicbreak}{\textbf{break}}
		\newcommand{\BREAK}{\STATE \algorithmicbreak}
		\REQUIRE Evolved block $b$, the training set $d$, the maximum number of times to stack $s_{max}$;
		\STATE $d_{train}, d_{test}\leftarrow$ Randomly split $d$ into training part $d_{train}$ and test part $d_{test}$ by 80\% and 20\%;
		\STATE $c_{set}, t\leftarrow$ Empty set of candidates, 0 as the current number of times to stack; 
		\WHILE{$t<s_{max}$}
		\STATE $t\leftarrow$ $t+1$;
		\STATE $c_{set}\leftarrow$ Stack $b$ for $t$ times to generate a candidate and append it to the candidate set;
		\ENDWHILE
		\STATE Use Adam optimisation to train each candidate in $c_{set}$ on $d_{train}$ and evaluate it on $d_{test}$;
		\STATE $c_{best} \leftarrow$ Find the candidate with the best classification on $d_{test}$;
		\RETURN $c_{best}$;
	\end{algorithmic}
\end{algorithm}



\section{Experiment Design}\label{SSS:effpnet_experiment}

In this section, the detailed design of the experiments will be depicted. The three benchmark datasets --- CIFAR-10, CIFAR-100 and SVHN, and the selected peer competitors will be discussed in Sections \ref{SSS:effpnet_dataset} and \ref{SSS:effpnet_competitors}. In addition, the parameter settings to run the experiments will be listed and explained in Section \ref{SSS:effpnet_parameter}. 

\subsection{Benchmark Datasets}\label{SSS:effpnet_dataset}

First of all, the dataset taken by the proposed method to generate the surrogate dataset needs to be selected. CIFAR-10 \cite{krizhevsky2009learning} can fit this purpose well because it is a widely-used benchmark dataset to evaluate image classification, and it is a medium-scale dataset of 10 classes comprised of various images with decent complexity. Therefore, there are three benefits of using CIFAR-10 to generate the surrogate dataset. The first benefit is that the surrogate dataset sampled from CIFAR-10 can reflect various images. The second is that the training process of CNNs on the surrogate dataset does not take too much computational resource. The third is that the reduced number of examples in the surrogate dataset does not make the classification task too hard during the fitness evaluation due to the small number of classes. Conversely, if CIFAR-100 is chosen instead of CIFAR-10, the surrogate dataset might not be sufficient to train an effective CNN to distinguish the images from 100 classes. There are 60,000 coloured images of 10 classes in the CIFAR-10 dataset, which contains 50,000 training images and 10,000 test images.  In addition, to assess the classification performance of the block, CIFAR-10 is used again to validate the effectiveness of the evolved block learned from a subset of itself. Last but not least, the transferability of the evolved block needs to be evaluated on other benchmark datasets, which are not seen by the proposed method during the evolutionary process. CIFAR-100 \cite{krizhevsky2009learning} is chosen because the image domain and the total number of images in CIFAR-100 are similar to those of CIFAR-10, but its number of classes is extended to 100, which results in a much more complex classification task. Besides CIFAR-100, the SVHN dataset \cite{netzer2011reading}, which is comprised of digit images from 0 to 9 obtained from house numbers in Google Street View images, could test the transferability of the evolved block in a different domain because its images are disparate from those of CIFAR-10. Hence, the evolved transferable block can be thoroughly verified in a similar domain and a different domain.

\subsection{Peer Competitors}\label{SSS:effpnet_competitors}

The peer competitors are selected mainly based on the availability of the performance reported on the above benchmark datasets and their relevance to the proposed method. Firstly, the performance of the evolved block on CIFAR-10, where the block is learned from its own surrogate dataset, needs to be assessed. There are three sets of competitors chosen for the comparison. The first set consists of a couple of the state-of-the-art CNNs designed manually, which are ResNet \cite{he2016deep} and DenseNet \cite{huang2017densely}. The second set incorporates automatically-designed CNNs by reinforcement learning, which are PNASNet \cite{liu2018progressive}, BockQNN \cite{zhong2018practical}, EAS \cite{cai2018efficient}, NASNet-A (7 @ 2304) \cite{zoph2018learning}, NASH (ensemble across runs) \cite{elsken2017simple} and NAS v3 max pooling \cite{zoph2016neural}. The last set is comprised of CNNs designed automatically by EC methods, which are EIGEN \cite{ren2019eigen}, RENAS \cite{chen2019renas}, AECNN \cite{sun2019completely}, AmoebaNet-B (6,128) \cite{real2018regularized}, Hier. repr-n, evolution (7000 samples), CGP-CNN(ResSet) \cite{suganuma2017genetic}, DENSER \cite{assunccao2018evolving} GeNet from WRN \cite{xie2017genetic}, CoDeapNEAT \cite{miikkulainen2019evolving} and LS-Evolution \cite{real2017large}. Since the proposed method is for automatically designing CNNs based on EC, the relevance of peer competitors increases from the first to the last set. Therefore, the number of competitors grows from the first to the last set as well. Furthermore, the performance of the evolved block transferred to the other two datasets needs to be verified. Based on the availability of the reported performance on CIFAR-100 and SVHN, the following peer competitors will be compared with the proposed method --- Network in Network \cite{lin2013network}, CiCNet \cite{pang2017convolution}, Deeply Supervised Net \cite{lee2015deeply}, FractalNet \cite{larsson2016fractalnet}, WideResNet \cite{zagoruyko2016wide}, ResNet \cite{huang2016deep} and DenseNet(k=12) \cite{huang2017densely}. 

\subsection{Parameter Settings}\label{SSS:effpnet_parameter}

\begin{table}[t]
	\renewcommand{\arraystretch}{1.3}
	\caption{Parameter settings}
	\label{table:effpnet_parameters}
	\centering
	\begin{tabular}{|c|c|}
		\hline
		Parameter & Value\\
		\hline
		\multicolumn{2}{|c|}{\textbf{EffPNet hyper-parameters}} \\
		\hline
		maximum number of layers in Section \ref{SSS:effpnet_encoding} & 16\\
		\hline
		range of growth rate in Section \ref{SSS:effpnet_encoding} & [11, 32]\\
		\hline
		threshold to activate surrogate in Section \ref{SSS:effpnet_surrogate_model} & 90\%\\
		\hline
		feature-cutting epochs \ref{SSS:effpnet_surrogate_model} & 10\\
		\hline
		data reduction ratio in \ref{SSS:effpnet_surrogate_dataset} & 10\%\\
		\hline
		downsampling factor in \ref{SSS:effpnet_surrogate_dataset} & 2\\
		\hline
		the maximum number of times to stack the block in \ref{SSS:effpnet_stack} & 5\\
		\hline
		\toprule
		\hline
		\multicolumn{2}{|c|}{\textbf{PSO parameters}} \\
		\hline
		inertia weight $w$ & 0.7298\\
		\hline
		acceleration coefficient $c1$ & 1.49618\\
		\hline
		acceleration coefficient $c2$ & 1.49618\\
		\hline
		velocity range & [-10.5, 10.5]\\
		\hline
		population size & 30\\
		\hline
		number of generations & 50\\
		\hline
	\end{tabular}
\end{table}

The hyper-parameters of the experiments are defined and listed in Table \ref{table:effpnet_parameters}. Firstly, the maximum number of layers is set to 16 and the range of growth rate for the block to be evolved is set to [11, 32] due to the hardware limit of running the experiments. Since the experiments run on a distributed system designed for evolving CNNs \cite{wang2019evolving}, the above two hyper-parameters for the block are designed based on the lowest GPU card - GeForce RTX 2070. Secondly, a set of hyper-parameters for the surrogate model and the surrogate dataset are designed. A high threshold of activating the surrogate model is set to 90\% accuracy, and the feature-cutting epochs of 10 is designed for the surrogate model to achieve high accuracy. The data reduction ratio and downsampling factor are 10\% and 2 to achieve a smaller surrogate dataset in terms of both the number of examples and the image size. Thirdly, the maximum number of times to stack the evolved CNNs is set to 5 based on the complexity of the benchmark datasets. Finally, the PSO parameters are designed according to the community convention \cite{shi1998parameter} \cite{trelea2003particle} \cite{van2006study}.

\section{Results and Analysis}\label{SSS:effpnet_result}

The section will endeavour to analyse the results from the experiments. For the result analysis, first of all, the performance on the three benchmark datasets --- CIFAR10, CIFAR100 and SVHN will be shown and compared with peer competitors. Then, the convergence of the proposed method will be visualised and analysed. Furthermore, the experimental results specifically related to the surrogate model will be analysed to demonstrate the effectiveness and efficiency of the surrogate model. In the end, the growth rates for different layers in the evolved blocks will be analysed to show the preference of growth rates for different layers in the evolved blocks. 

\subsection{Performance Comparisons}\label{SSS:effpnet_accuracy}

\subsubsection{Performance Comparisons on CIFAR-10}

\begin{table}[t]
	\renewcommand{\arraystretch}{1.3}
	\caption{Performance comparison with peer competitors on CIFAR-10}
	\label{table:epsocnn_performance}
	\centering
	\begin{tabular}{|P{0.2\linewidth}|P{0.2\linewidth}|P{0.2\linewidth}|P{0.2\linewidth}|}
		\hline
		Method & Error rate\% & Number of Parameters & Computational Cost\\
		\hline
		ResNet-110 \cite{he2016deep} & 6.43 & \textbf{1.7M} & --\\
		\hline
		DenseNet(k = 40) \cite{huang2017densely} & 3.74 & 27.2M & --\\
		\hline
		\toprule
		\hline
		PNASNet \cite{liu2018progressive} & \textbf{3.41 $\pm$ 0.09} & 3.2M & 225 GPU-days\\
		\hline
		BockQNN \cite{zhong2018practical} & 3.54 & 39.8M & 96 GPU-days\\
		\hline
		EAS \cite{cai2018efficient} & 4.23 & 23.4M & $<$10 GPU-days\\
		\hline
		NASNet-A (7 @ 2304) \cite{zoph2018learning} & \textbf{2.97} & 27.6M & 2,000 GPU-days\\
		\hline
		NASH (ensemble across runs) \cite{elsken2017simple} & 4.40 & 88M & 4 GPU-days\\
		\hline
		NAS v3 max pooling \cite{zoph2016neural} & 4.47 & 7.1M & 22,400 GPU-days\\
		\hline
		\toprule
		\hline
		EIGEN \cite{ren2019eigen} & 5.4 & \textit{2.6M} & \textbf{2 GPU-days}\\
		\hline
		RENAS \cite{chen2019renas} & \textbf{2.88} & 3.5M & 6 GPU-days\\
		\hline
		AECNN \cite{sun2019completely} & 4.3 & \textbf{2.0M} & 27 GPU-days\\
		\hline
		AmoebaNet-B (6,128) \cite{real2018regularized} & \textbf{2.98} & 34.9M & 3150 GPU-days\\
		\hline
		Hier. repr-n, evolution (7000 samples) \cite{liu2017hierarchical} & 3.75 & -- & 300 GPU-days\\
		\hline
		CGP-CNN(ResSet) \cite{suganuma2017genetic} & 5.98 & \textbf{1.68M} & 29.8 GPU-days\\
		\hline
		DENSER \cite{assunccao2018evolving} & 5.87 & 10.81M & --\\
		\hline 
		GeNet from WRN \cite{xie2017genetic} & 5.39 & -- & 100 GPU-days\\
		\hline
		CoDeapNEAT \cite{miikkulainen2019evolving} & 7.3 & -- & --\\
		\hline
		LS-Evolution \cite{real2017large} & 4.4 & 40.4M & $>$2,730 GPU-days\\
		\hline
		\toprule
		\hline
		\textbf{EffPnet (Best classification accuracy)} & 3.49 & 2.54M & $<$3 GPU-days\\
		\hline
		\textbf{EffPnet (10 runs)} & 3.576$\pm$0.0078 & 2.68M$\pm$0.015M & $<$3 GPU-days\\
		\hline
	\end{tabular}
\end{table}

Table \ref{table:epsocnn_performance} lists the classification accuracy, number of parameters and time taken to obtain the final CNNs of the proposed method and the selected peer competitors. In the columns of the \textit{Error rate\%} and the \textit{Number of Parameters}, if the result is obtained from multiple runs, the mean value and the standard deviation will be written as $mean\;value\pm standard\;deviation$. For the classification accuracy, comparing the best classification accuracy achieved by the proposed method with those of the 18 peer competitors, the proposed method achieved the 5th best, i.e. underperforming four peer competitors, whose error rates are in bold font in the table, with a very small margin. By applying Mann-Whitney-Wilcoxon (MWW) statistical test on the classification accuracies from the 10 runs of the proposed method and the error rate of 3.74\% from DenseNet, it shows that the proposed method is significantly better than the error rate of DenseNet. With regard to the number of parameters, the smallest CNN found by the proposed method has slightly more parameters than 3 of the 18 competitors. 
However, the three competitors with competitive sizes perform much worse than the proposed method in terms of classification accuracy. In respect of the computational cost, only EIGEN took less time than the proposed method by 1 GPU-day, but its classification accuracy is almost 2\% worse than that of the proposed method. Overall, the proposed method demonstrates its strong competitiveness with regard to the performance of all three measurements compared to the 18 peer competitors --- the 5th best in the error rate, the 4th best in the number of parameters and the 2nd best in the computational cost. 

\subsubsection{Transferability on CIFAR-100 and SVHN}
\begin{table}[ht]
	\renewcommand{\arraystretch}{1.3}
	\caption{Error rate comparison with peer competitors on CIFAR-100 and SVHN}
	\label{table:epsocnn_performance_tl}
	\centering
	\begin{tabular}{|P{0.4\linewidth}|P{0.2\linewidth}|P{0.2\linewidth}|}
		\hline
		Method & CIFAR-100 & SVHN\\
		\hline
		\hline
		Network in Network \cite{lin2013network} & 35.68 & 2.35 \\
		\hline
		Deeply Supervised Net \cite{lee2015deeply} & 34.57 & 1.92 \\
		\hline
		CiCNet \cite{pang2017convolution} & 24.82 & -- \\
		\hline
		FractalNet \cite{larsson2016fractalnet} & 23.30 & 2.01 \\
		\hline
		Wide ResNet \cite{zagoruyko2016wide} & 22.07 & \textit{1.85} \\
		\hline
		ResNet \cite{huang2016deep} & 27.22 & 2.01 \\
		\hline
		DenseNet(k=12) \cite{huang2017densely} & 20.20 & \textbf{1.67} \\
		\hline
		\hline
		\textbf{EffPNet (Best)} & 18.49 & 1.82 \\
		\hline
		\textbf{EffPNet (10 runs)} & 18.70$\pm$0.1620 & 1.85$\pm$0.0273 \\
		\hline
	\end{tabular}
\end{table}

To verify the effectiveness of the block's transferability, the transferable block learned from CIFAR-10 is stacked and the best CNN architecture is selected based on the CIFAR-100 and SVHN datasets, respectively, by following the method described in Section \ref{SSS:effpnet_stack}. The classification error rates on CIFAR-100 and SVHN are listed in Table \ref{table:epsocnn_performance_tl} along with those of its peer competitors. It can be observed that on the CIFAR-100 dataset, the proposed method statistically significantly outperforms all the compared methods according to the MWW test. For SVHN, the best error rate achieved by the proposed method exceeds the peer competitors apart from DenseNet. However, after applying the MWW statistical test, the proposed method is not significantly better than Wide ResNet, so it shares the second place on SVHN. To sum up, the proposed method has demonstrated its transferability by achieving very competitive classification accuracy in two different benchmark datasets. 

\subsection{Convergence Analysis}\label{SSS:effpnet_convergence}


\begin{figure}[ht]
	\centering
	\begin{subfloat}[]{
		\includegraphics[width=0.22\textwidth]{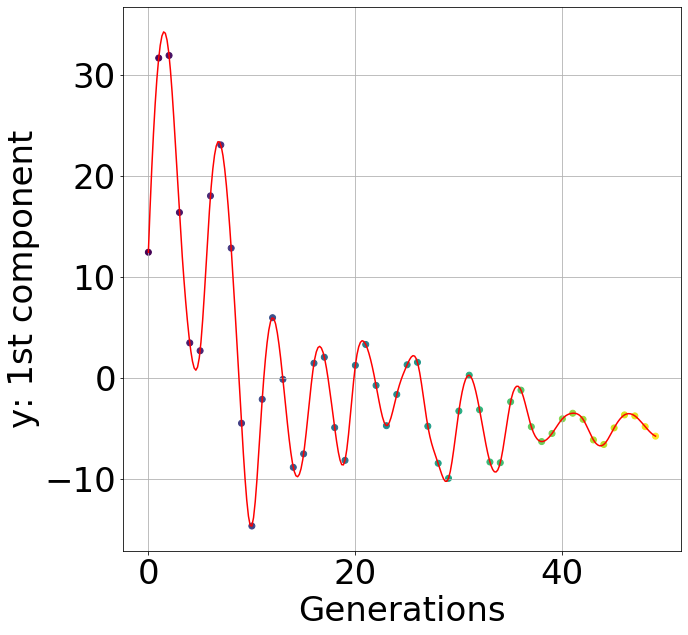}
		\label{fig:ec-convergence-particle-position-1}
	}
	\end{subfloat}
	\begin{subfloat}[]{
		\includegraphics[width=0.22\textwidth]{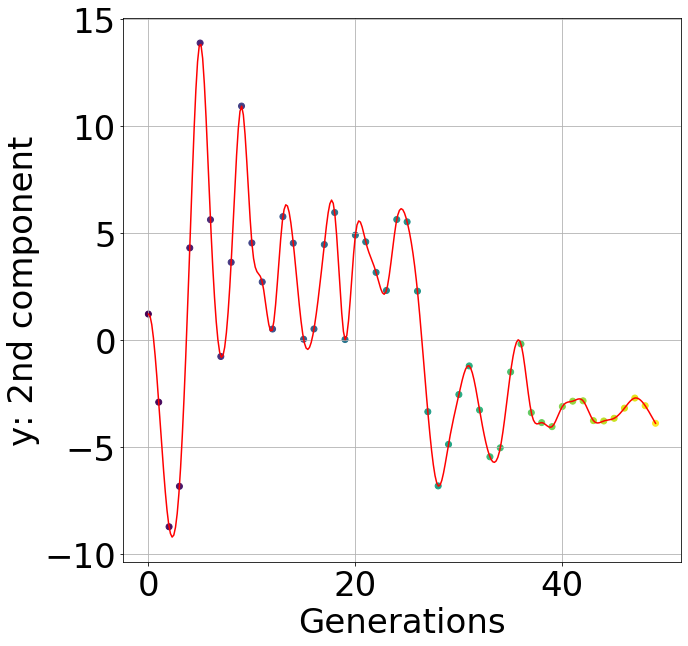}
		\label{fig:ec-convergence-particle-position-2}
	}
	\end{subfloat}
	\caption{Position convergence analysis. (a). 1st component of converged particle. (b). 2nd component of converged particle.}
	\label{fig:ec-convergence-particle-position}
\end{figure}

The convergence of the particle's position is investigated. An example of the particle's position convergence is shown in Fig. \ref{fig:ec-convergence-particle-position-1} and Fig. \ref{fig:ec-convergence-particle-position-2}. Two principal components are extracted by applying principal component analysis (PCA) \cite{abdi2010principal} \cite{wold1987principal} on the data comprised of the position vectors of all the particles. The movements of the particle's position during the whole evolutionary process are drawn based on the first and second principal components, respectively. It is clear that the particle fluctuates at the beginning, but the fluctuation becomes smaller and smaller for both of the principal components. In the end, the fluctuation curve is almost flattened, which demonstrates that the particle's position converges gradually during the evolutionary process.

\subsection{Analysis on Surrogate-assistance}\label{SSS:effpnet_surrogate_analysis}

\subsubsection{Surrogate Model Performance}\label{SSS:effpnet_surrogate_model_analysis_performace}

\begin{figure}[ht]
	\centering 
	\includegraphics[width=0.8\linewidth]{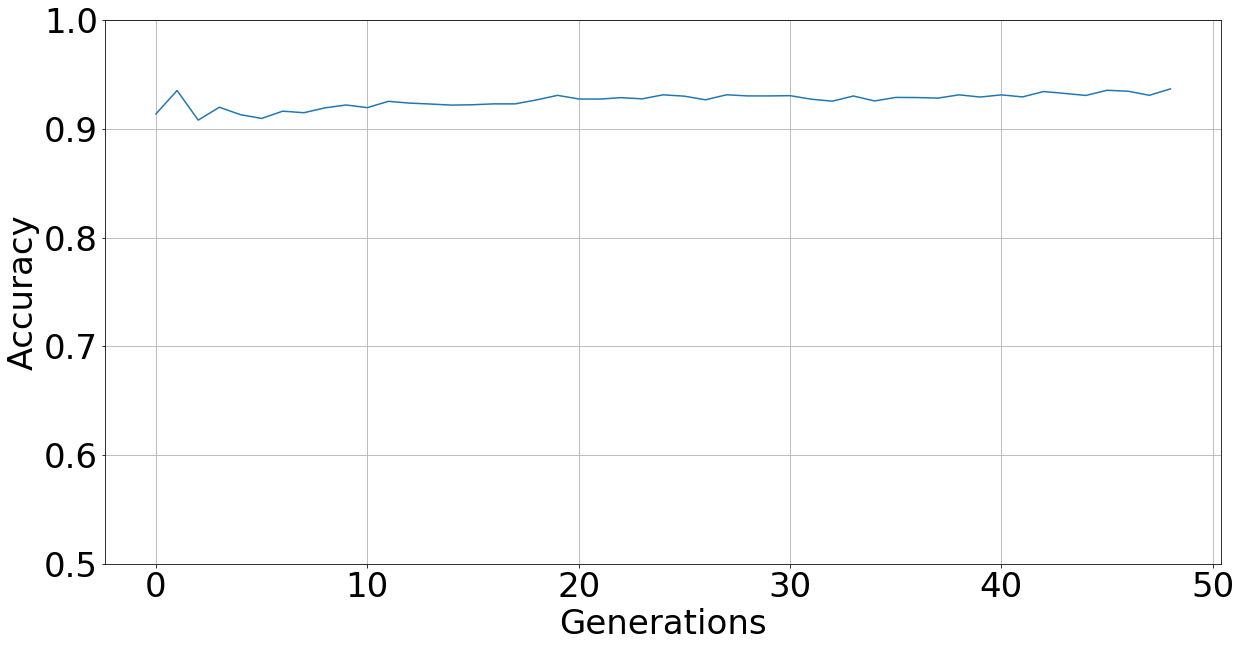}
	\caption{Surrogate model accuracies across generations. The statistical details are as follows --- the mean accuracy of 92.63\%, the standard deviation of 0.66\%, the minimum accuracy of 90.81\% and the maximum accuracy of 93.68\%.}
	\label{fig:effpent_svc_accuracies}
\end{figure}

Section \ref{SSS:effpnet_surrogate_model} describes that the task of predicting the performance of CNNs has been transformed into a binary classification task. The performance of the surrogate model can be measured by the classification accuracy, which reflects how accurate the predictions of the surrogate model are. The promising performance is supported by plotting the classification accuracy of the surrogate model across the generations of the evolutionary process as shown in Fig. \ref{fig:effpent_svc_accuracies}. The high accuracy of more than 90\% demonstrates that the surrogate model can distinguish the better CNN between a pair of given CNNs, which can efficiently and effectively assist the fitness evaluation. The very small standard deviation shows that the surrogate model consistently performs well across all of the generations, which proves its great stability during the whole evolutionary process. Therefore, the surrogate model can assist the fitness evaluation effectively during the entire search process. Another observation is that the surrogate model performs well even at the first generation with the constructed dataset from training only 30 blocks, which indicates a small number of training examples can fit the SVM classifier in the surrogate model reasonably well. 

\begin{figure*}[ht]
	\centering 
	\begin{subfloat}[]{
		\includegraphics[width=0.7\textwidth]{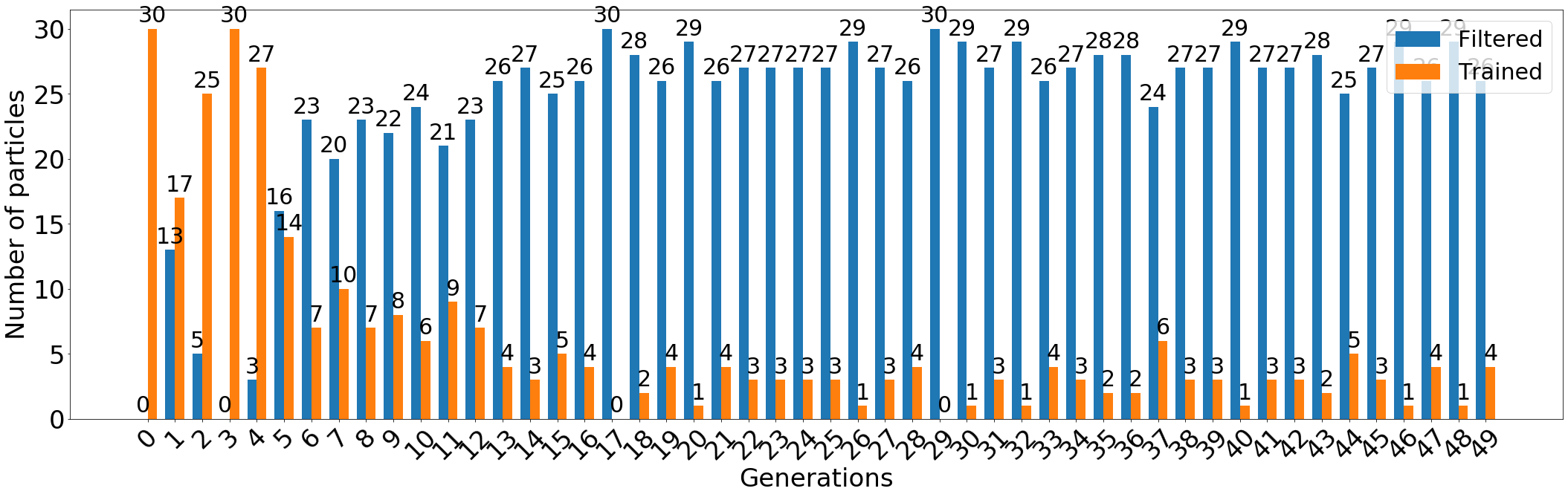}
		\label{fig:surrogate-svc-stats-gens}
	}
	\end{subfloat}
	\begin{subfloat}[]{
		\includegraphics[width=0.25\textwidth]{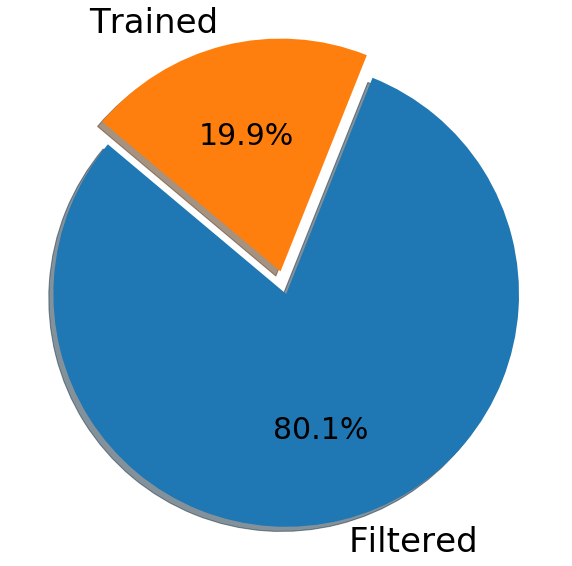}
		\label{fig:surrogate-svc-stats-sum}
	}
	\end{subfloat}
	\caption{Surrogate-assisted fitness evaluation stats. (a). Surrogate-assisted fitness evaluation across generations. (b). Surrogate-assisted fitness evaluation summary.}
\end{figure*}

Besides the accuracy, another key metric of assessing the surrogate model is the number of particles filtered by the surrogate model. Before discussing the experimental results in this metric, two terms need to be defined first. Firstly, a \textit{filtered} particle is defined as the particle predicted by the surrogate model to underperform the particle's best-so-far position, which does not need to undergo the time-consuming CNN training process. The \textit{trained} particle means the particle predicted to outperform its best-so-far position, which, therefore, has to be accessed by training the CNN. Fig. \ref{fig:surrogate-svc-stats-gens} shows the numbers of the filtered and trained particles, respectively, at each generation. It can be seen that the number of trained particles exceeds the number of filtered particles only at the first several generations, but for the majority of the generations, the filtered particles significantly outnumber the trained particles. Fig. \ref{fig:surrogate-svc-stats-sum} illustrates the percentages of filtered and trained particles, respectively, in total, which shows that 80.1\% of the particles have been filtered by the surrogate model. Therefore, the surrogate model has successfully sped up the fitness evaluation by preventing 80.1\% of the particles from being trained by the time-consuming CNN training process. 


Another simple approach of accelerating the fitness evaluation is to train the CNNs represented by the particles for a small number of epochs, e.g. 10 epochs, and use the accuracy at the 10th epoch as the fitness value \cite{wang2018evolving} \cite{wang2019hybrid}. Based on the above-constructed dataset, an evaluation of using the accuracy at the 10th epoch as the sole indicator of the CNN's performance is done, which only achieves an accuracy of 69.94\%. Considering that this is a binary classification problem, the method of solely utilising the accuracy of the 10th epoch may be able to indicate the final classification accuracy, but the accuracy is not good. Instead, the surrogate model has consistently achieved an accuracy of more than 90\% during the whole evolutionary process, so the surrogate-assisted fitness evaluation can be deemed as a reliable approach to speed up the fitness evaluation process. 

\subsubsection{Surrogate Model with Different Feature Combinations}\label{SSS:effpnet_surrogate_model_analysis_features}

\begin{table}[ht]
	\renewcommand{\arraystretch}{1.3}
	\caption{Surrogate model accuracies with various combinations of features}
	\label{table:effpnet_surrogate_accuracy_features}
	\centering
	\begin{tabular}{|l|P{0.2\linewidth}|}
		\hline
		Features & Accuracy \\
		\hline
		losses & 82.02\% \\
		\hline
		accuracies & 86.27\% \\
		\hline
		block parameters & 70.60\% \\
		\hline
		losses + accuracies & 86.96\% \\
		\hline
		losses + accuracies + block parameters & 91.13\% \\
		\hline
	\end{tabular}
\end{table}

To analyse the features of the data used to train the surrogate model, the performances of the surrogate model with various combinations of features are evaluated and listed in Table \ref{table:effpnet_surrogate_accuracy_features}. A few interesting points can be observed from the table. Firstly, the surrogate model with only the first 10 losses or only the first 10 accuracies achieves more than 10\% accuracy than only using the block parameters. This indicates that the losses and accuracies are more important features than the block parameters. 
Secondly, despite good accuracies achieved by using the losses and accuracies individually, when combining the losses and accuracies as the features for the surrogate model, the improvement is tiny with only 0.69\% more accurate than solely taking the accuracies as the features. It implies that these two sets of features may be redundant. 
At last, it can be found that there is a decent improvement of the accuracy by combining the block parameters with the losses and accuracies, which is almost 4.17\% more accurate than using the features without the block parameters. Since the combination of using all features shows the best performance, the surrogate model in the proposed method has chosen the proper features in terms of achieving the best classification accuracy.

\subsubsection{Data Analysis of Block Training History}\label{SSS:effpnet_fitness_db_analysis}

\begin{figure*}[ht]
	\centering 
	\begin{subfloat}[PC1 of losses.]{
			\includegraphics[width=0.30\textwidth]{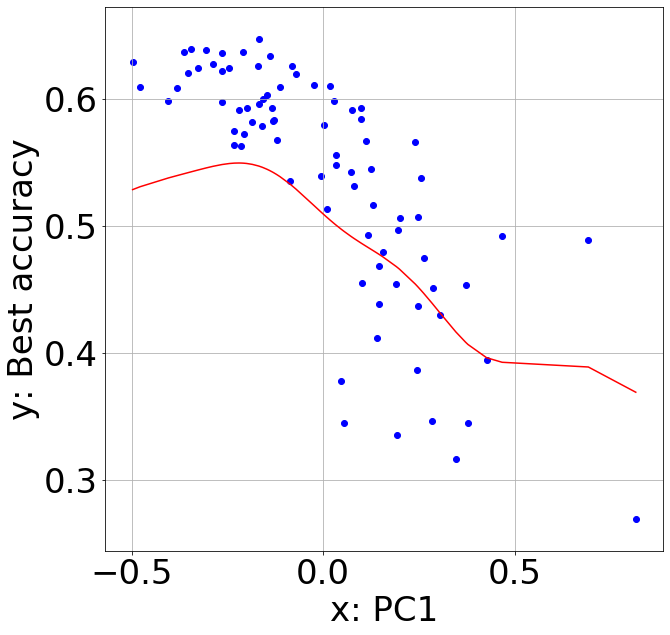}
			\label{fig:effpent_fitness_db_analysis_11}
		}
	\end{subfloat}
	\begin{subfloat}[PC1 of accuracies.]{
			\includegraphics[width=0.30\textwidth]{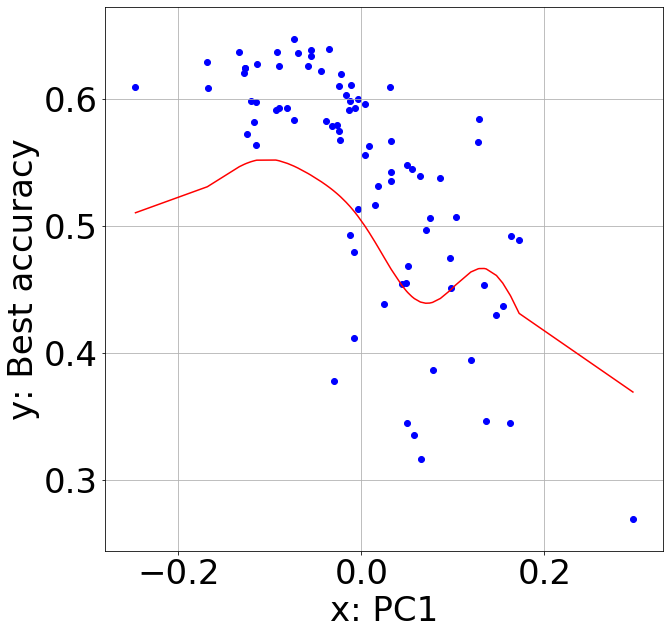}
			\label{fig:effpent_fitness_db_analysis_21}
		}
	\end{subfloat}
	\begin{subfloat}[PC1 of block vectors.]{
			\includegraphics[width=0.30\textwidth]{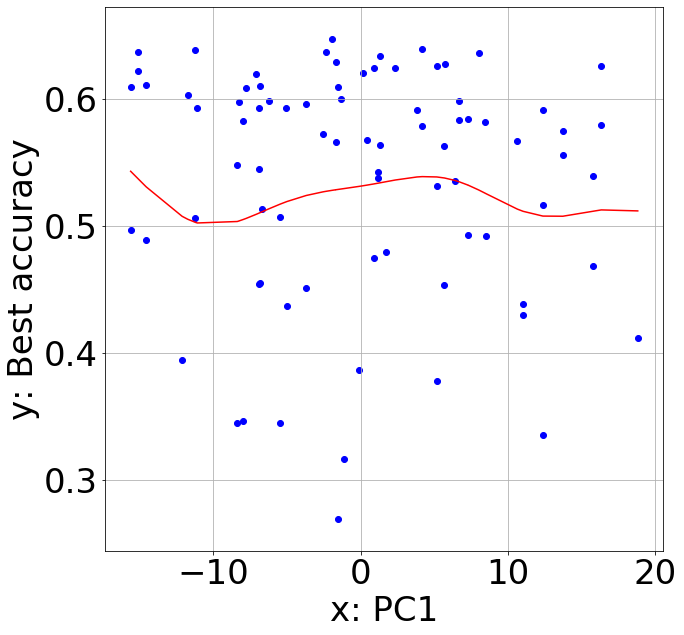}
			\label{fig:effpent_fitness_db_analysis_31}
		}
	\end{subfloat}
	\caption{The x-axis is the PC1 and the y-axis is the best classification accuracy. The red line is the pattern discovered by the SVR model. (a). Extract the first 10 losses from the \textit{block training history} and apply PCA on the extracted data to achieve the PC1. (b). Extract the first 10 accuracies from the \textit{block training history} and apply PCA on the extracted data to achieve the PC1. (c). Extract the 16 growth rates from the \textit{block training history} and apply PCA on the extracted data to achieve the PC1.}
	\label{fig:effpent_fitness_db_analysis}
\end{figure*}

To further obtain an insight on how the features used for the surrogate model affect the classification accuracy of the dense block, the parameters of the evolved block, the losses and accuracies of the first 10 epochs, and the best accuracy during the whole training process are extracted. Therefore, it is feasible to discover the patterns of how the best classification accuracy is influenced by the block parameters, and the performance trend including the losses and the accuracies of the first 10 epochs. The best classification accuracy is the dependent variable and the remaining features are the explanatory variables. The most straightforward method of analysing data is to visualise them, so the dimensionality of the explanatory variables needs to be reduced to 2-D or 3-D. As the purpose of the analysis is to find the importance of the explanatory variables, the variance is important for keeping the feature information during the dimensionality reduction. Thus, PCA \cite{abdi2010principal} \cite{wold1987principal} is adopted because it is the most popular multivariate statistical technique, which was designed to reduce the dimensionality by extracting the most important information. Once the visualisation is done, it would be more explainable to detect the pattern of the data points. A simple support vector regression (SVR) method \cite{smola2004tutorial} \cite{chang2011libsvm} is chosen to discover the pattern, i.e. in Fig. \ref{fig:effpent_fitness_db_analysis}, an SVR model is fit by the blue points in each of the sub-figures and the trained SVR model draws the red line as the discovered pattern. 

Fig. \ref{fig:effpent_fitness_db_analysis} plots the first principal component (PC1) versus the classification accuracy. It can be observed from Sub-figure \ref{fig:effpent_fitness_db_analysis_11}, a general pattern for the PC1 of the first 10 losses is presented because the blue points gather near the red line. A kind of correlation between x-axis and y-axis can be observed, where the points closer to the left side of the x-axis tend to achieve higher y values. By examining Sub-figure \ref{fig:effpent_fitness_db_analysis_21}, similarly, a general pattern of the PC1 is also detected. It can be seen that the similar patterns are found between the sub-figures \ref{fig:effpent_fitness_db_analysis_11} and \ref{fig:effpent_fitness_db_analysis_21}, which makes sense because the accuracy increases while the training losses decreases during the neural network training process by SGD algorithms \cite{sutskever2013importance} \cite{kingma2014adam}, especially at the early stage without over-fitting issues. Sub-figures \ref{fig:effpent_fitness_db_analysis_31} does not show a clear pattern for the PC1 because the red line cannot match the data points well. To summarise, the performance trend carried by the losses and accuracies from the first 10 epochs are crucial features to predict the final performance of the dense block; while the block parameters are less important features, but they may be used as assistant features to improve the accuracy of predicting the final performance. This is also consistent with the performance of the surrogate model with various feature combinations discussed in Section \ref{SSS:effpnet_surrogate_model_analysis_features}, where it showed the losses and the accuracies were more important features than the block parameters.

\subsection{Growth Rate Analysis}\label{SSS:effpnet_architecture}

\begin{figure*}[ht]
	\begin{subfloat}[Accuracy.]{
		\includegraphics[width=0.18\textwidth]{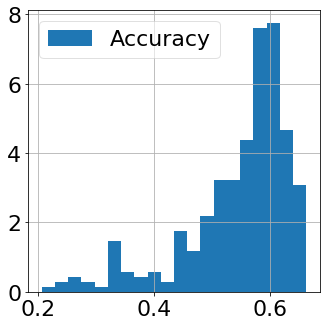}
		\label{fig:effpnet-growth-rate-accuracy-all}
	}
	\end{subfloat}
	\begin{subfloat}[Layer-1 growth rate.]{
		\includegraphics[width=0.18\textwidth]{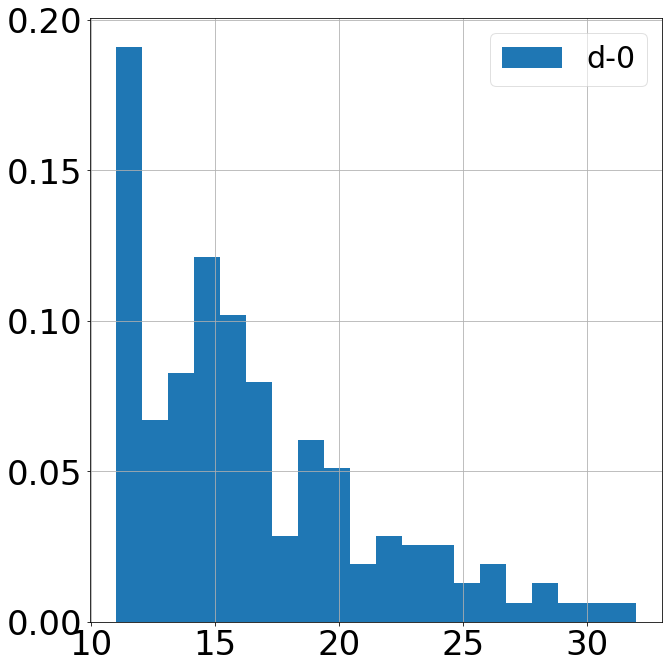}
		\label{fig:effpnet-growth-rate-distribution-d0}
	}
	\end{subfloat}
	\begin{subfloat}[Layer-9 growth rate.]{
		\includegraphics[width=0.18\textwidth]{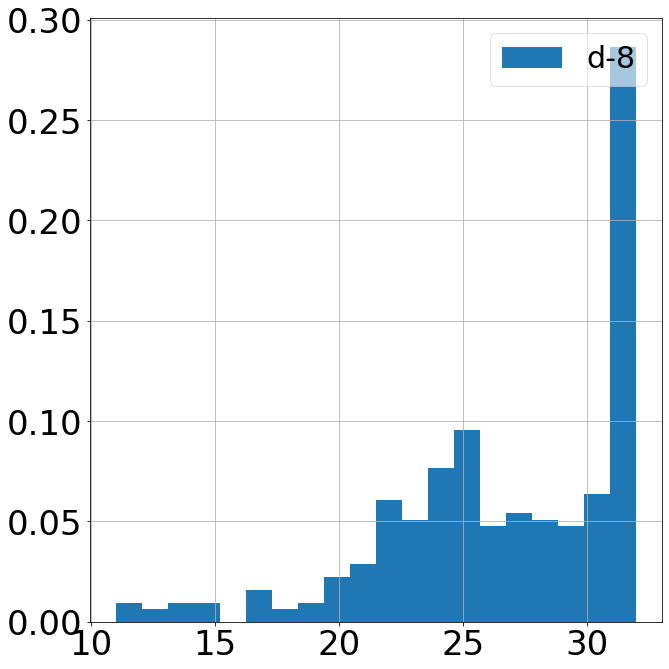}
		\label{fig:effpnet-growth-rate-distribution-d8}
	}
	\end{subfloat}
	\begin{subfloat}[Layer-10 growth rate.]{
		\includegraphics[width=0.18\textwidth]{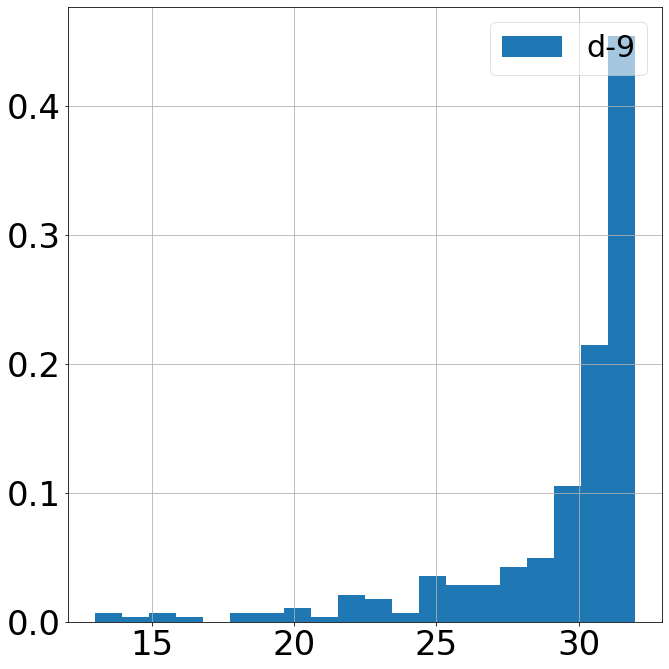}
		\label{fig:effpnet-growth-rate-distribution-d9}
	}
	\end{subfloat}
	\begin{subfloat}[Layer-15 growth rate.]{
		\includegraphics[width=0.18\textwidth]{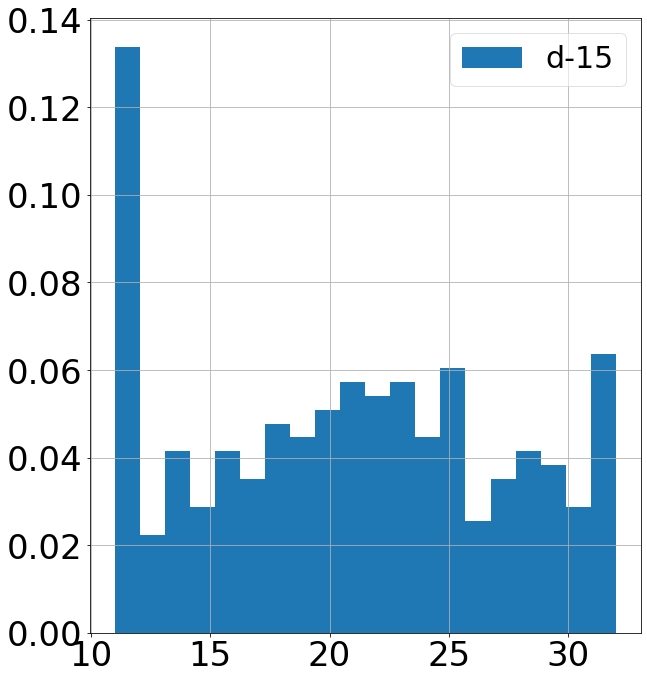}
		\label{fig:effpnet-growth-rate-distribution-d15}
	}
	\end{subfloat}
	\caption{Distribution of accuracy and growth rates. All of the sub-figures are histogram plots with 20 bins. (a). x-axis is the accuracy and y-axis is the number of evaluated blocks, whose accuracies fall into the range of the accuracy bin. (b)(c)(d)(e). x-axis is the growth rate and y-axis and y-axis is the number of evaluated blocks, whose growth rates fall into the growth rate bin.}
	\label{fig:effpnet-distribution-analysis}
\end{figure*}

\begin{figure*}[ht]
	\centering 
	\includegraphics[width=0.96\textwidth]{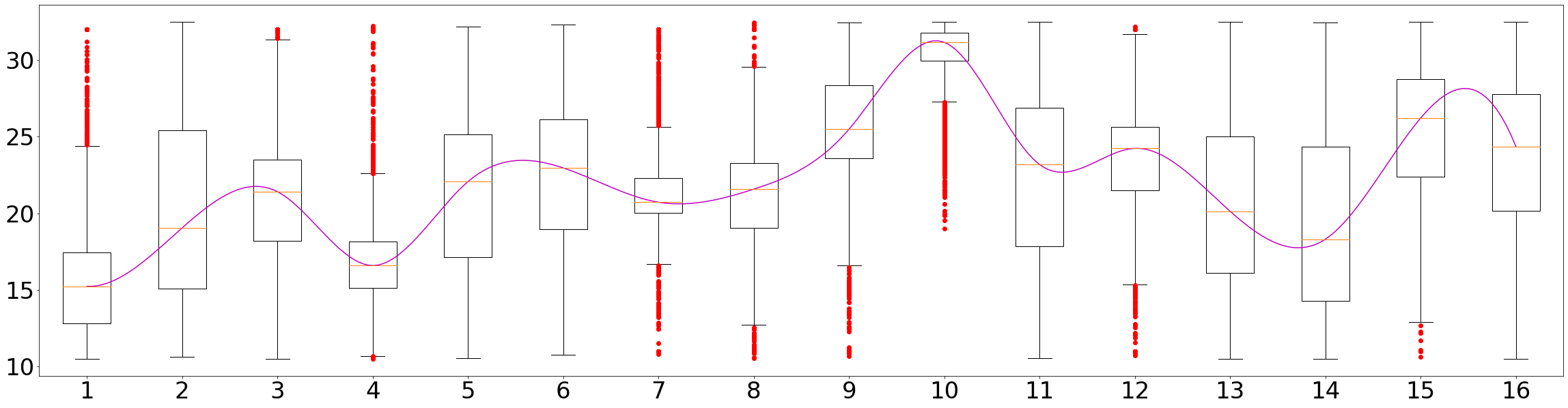}
	\caption{Statistics of the growth rate in total for each layer.}
	\label{fig:effpnet-growth-rate-dimension-all}
\end{figure*}


To explore the impact of various growth rates for each layer, the distribution of the accuracy and the growth rates of the evaluated blocks represented by the particles is drawn in the sub-figures of Fig. \ref{fig:effpnet-distribution-analysis}. Fig. \ref{fig:effpnet-growth-rate-accuracy-all} outlines the accuracy distribution of the evaluated blocks. It is obvious that the area with higher accuracies are explored much more than other space, especially the search space where the accuracy is around 0.6. This indicates that the blocks with higher accuracies are well explored by the proposed method. Looking at the distribution of the growth rates for the first and last layers, respectively, as shown in Fig. \ref{fig:effpnet-growth-rate-distribution-d0} and \ref{fig:effpnet-growth-rate-distribution-d15}, the bar with the smallest growth rate is longer than others; while, for the two middle layers shown in Fig. \ref{fig:effpnet-growth-rate-distribution-d8} and \ref{fig:effpnet-growth-rate-distribution-d9}, the largest growth rate significantly outnumbers others. Therefore, for the blocks with higher accuracies dominating the distribution, different layers tend to prefer different growth rates. In addition to the distribution, Fig. \ref{fig:effpnet-growth-rate-dimension-all} is a box-plot of the growth rates for each layer of all of the evaluated blocks with a curved line connecting the median values of the growth rates. From the shape of the curved line, different growth rates have been chosen for different layers. The first layer has the smallest median value and the 10th layer obtains the largest median value, which implies that the middle layer needs a larger growth rate than the other layers towards the first or the last layer. To conclude, the proposed method has searched the areas of the search space, where the classification accuracy tends to be better, and it is meaningful to optimise the growth rates for each layer in the dense blocks instead of using a fixed growth rate for all as proposed in the original DenseNet paper.

\section{Conclusions and Future Work}\label{SSS:effpnet_conclusion}

In conclusion, a new surrogate-assisted PSO method has been proposed to effectively and efficiently evolve CNN blocks that are transferable to different domains in an automatic manner. This is supported by the experimental results, which have demonstrated that the proposed method is able to efficiently learn an effective block from the CIFAR-10 dataset by achieving promising performance on CIFAR-10. Furthermore, the evolved blocks have exhibited their transferability by achieving competitive classification accuracies on the CIFAR-100 and SVHN datasets. To achieve the promising performance in terms of both classification accuracy and computational cost, firstly, a surrogate model and a surrogate dataset were proposed to significantly accelerate the fitness evaluations. The reliability of the surrogate model and the surrogate dataset has been upheld by visualising and analysing the accuracy and decision boundary of the surrogate model. Secondly, the surrogate model and surrogate dataset were integrated into the PSO algorithm to form a surrogate-assisted PSO to achieve the goal of efficiently searching for optimal blocks. Last but not least, an encoding strategy to accommodate various growth rates of different layers in variable-length blocks was proposed, and the growth rate of each layer in the block was analysed to show the necessity of having different growth rates at different layers of the block. Overall, the goal of this paper and the specific objectives have been fulfilled. 

This work demonstrates the potential of efficiently learning a dense block, which could be effectively transferred to other domains. However, the prior knowledge of DenseNet is utilised and the search space is restricted to search for hyper-parameters of a dense block. It would be interesting to explore CNN architectures in a more flexible search space without any prior knowledge. For example, the search space could be composed of only the basic convolutional and pooling layers and the connections in the block could be sparsely- or densely- connected. In addition, the proposed method used a feed-forward fashion to stack the evolved block at the last step, but the shortcut connections could also be used. By loosening the restrictions of the search space, the major advantage is to enable the possibility of discovering unknown CNN architectures, which could outperform the state-of-the-art CNNs. However, the biggest challenge is the efficiency of the search algorithms. Developing an effective and efficient method to explore CNN architectures in a large search space could be very attractive research in the future.


%




\ifCLASSOPTIONcaptionsoff
  \newpage
\fi



\bibliographystyle{IEEEtran}
\bibliography{effpnet}
%

%

\begin{IEEEbiography}{Bin Wang}
Biography text here.
\end{IEEEbiography}

\begin{IEEEbiography}{Bing Xue}
Biography text here.
\end{IEEEbiography}

\begin{IEEEbiography}{Mengjie Zhang}
	Biography text here.
\end{IEEEbiography}






\end{document}